\documentclass[
]{llncs}

\usepackage{relsize,etoolbox}%
\usepackage[utf8]{inputenc}
\usepackage{amsmath}
\usepackage{graphicx}
\usepackage{amsfonts}
\usepackage{amssymb}
\usepackage{url}

\AtBeginEnvironment{quote}{\smaller}

\begin{document}

{\let\thefootnote\relax\footnotetext{Copyright \textcopyright\ 2021 for this paper by its authors. Use permitted under Creative Commons License Attribution 4.0 International (CC BY 4.0). CLEF 2021, 21-24 September 2021, Bucharest, Romania.}}



\title{Accenture at CheckThat! 2021: Interesting claim identification and ranking with contextually sensitive lexical training data augmentation}


\author{Evan Williams\orcidID{0000-0002-0534-9450} \and Paul Rodrigues\orcidID{0000-0002-2151-636X} \and Sieu Tran\orcidID{0000-0003-0017-4329}}
\institute{
Accenture,
  800 N. Glebe Rd., Arlington, 22209, USA\\
\url{e.m.williams@accenture.com}\\
\url{paul.rodrigues@accenture.com}\\
\url{sieu.tran@accenture.com}\\
}

\maketitle

\begin{abstract}
  This paper discusses the approach used by the Accenture Team for CLEF2021 CheckThat! Lab, Task 1, to identify whether a claim made in social media would be interesting to a wide audience and should be fact-checked.  Twitter training and test data were provided in English, Arabic, Spanish, Turkish, and Bulgarian.  Claims were to be classified (check-worthy/not check-worthy) and ranked in priority order for the fact-checker.  Our method used deep neural network transformer models with contextually sensitive lexical augmentation applied on the supplied training datasets to create additional training samples.  This augmentation approach improved the performance for all languages.  Overall, our architecture and data augmentation pipeline produced the best submitted system for Arabic, and performance scales according to the quantity of provided training data for English, Spanish, Turkish, and Bulgarian.  This paper investigates the deep neural network architectures for each language as well as the provided data to examine why the approach worked so effectively for Arabic, and discusses additional data augmentation measures that should could be useful to this problem.  

\end{abstract}

\begin{keywords}
  fact-checking,
  claim retrieval, 
  check-worthy,
  BERT,
  RoBERTa
\end{keywords}


\section{Introduction}

Labeled data for some machine learning tasks can be quite rare and valuable.  Data labeling is a time consuming task, and can be  expensive if subject matter or language expertise are required.  Machine learning engineers know that generally, the larger the training set, the higher accuracy the classifier will have \cite{dumais1998inductive}, so they often request more data. 

Further, engineers have been taught to prefer balanced data sets to train more robust classifiers.  Most annotation processes yield unbalanced datasets naturally.  Data scientists often have four options: upsampling, downsampling, cost-sensitive learning, or active learning to seek a balanced dataset through additional annotation.  Downsampling reduces the data provided to the learner, which removes valuable labeled data from the dataset. Active learning incurs additional labeling cost.  Upsampling is an attractive alternative.  Upsampling in NLP application areas could involve exact text duplication or text augmentation.

The CLEF CheckThat! Labs provide shared training data for all the groups.  This data provided in 2020 and 2021 was naturally unbalanced, predominantly consisting of documents that are not check-worthy.  

Last year we published a paper at CLEF CheckThat! which used back translation to balance the classes in our Arabic training data.  \cite{williams:ea:20} The noise introduced by the machine translation system provided for an improvement in classifier performance.  This method resulted in the best performing Arabic model in the Lab.  For this year's Lab, we again endeavored to generate additional labeled data from the provided labeled data.  

This year we used a different technique, contextually sensitive lexical augmentation, and we applied the approach to all the languages.  Our technique uses BERT and RoBERTa models to replace text from the provided sample to construct alternative samples for the positive class tweets.  We used this as additional training input to our transformer neural networks.  








\subsection{CheckThat! Lab}
CLEF CheckThat! is a series of annual challenges to identify the best algorithms for automated fact-checking.  The 2021 Lab focused on social media and news articles.  \cite{clef-checkthat:2021:LNCS}
Accenture focused on Task 1 of this Lab, which required identification if a claim on Twitter was worth fact-checking, and ranking the claim for how check-worthy the claim was.  \cite{clef-checkthat:2021:task1} This challenge focused on Arabic, Bulgarian, English, Spanish, and Turkish.

Accenture's paper in the 2020 Lab reached 1st place in the English track, and 1st, 2nd, 3rd, and 4th in the Arabic track. \cite{williams:ea:20}. This year, only one submission per team per language was accepted for final reporting.

\subsection{Data Augmentation}

Data augmentation is considered an important component in Deep Learning workflows \cite{raghu2020survey}. These techniques are commonly applied in speech recognition (e.g. insertion of babble), and computer vision (e.g. image rotation) systems, but are not as commonly applied in natural language processing workflows due to differences in resources and techniques based on language and task. \cite{wei-zou-2019-eda}  

Augmentation in an NLP context can take many forms.  Words can be replaced with synonyms, antonyms, hypernyms, homonyms, or semantic neighbors.  Words can be deleted, or inserted either randomly or where the lexical insertion best fits a language model.  Words can be misspelled intentionally, either phonetically or at the character level, with a deletion, insertion, replacement or a swap of character sequence.  There are numerous methods for NLP data augmentation, but none are commonplace. 

\cite{wei-zou-2019-eda} explored the use of text augmentation at the lexical level on five classification tasks, including a subjective/objective discrimination task.  For each sentence in a training set, the researchers randomly selected between four operations-random lexical insertion, random lexical substitution, random lexical deletion, and synonym replacement.  They showed that by applying these random augmentations, they were able to boost accuracy on all five classification tasks.  Further, in an ablation study, the researchers showed that each of the four operations contributed to accuracy improvements.  They noted that their augmentation strategy was particularly beneficial to smaller datasets.  The authors performed their experiments on LSTM-RNN and CNN architectures.  We are not aware of any papers describing NLP augmentation as input to transformer architectures for document classification tasks.

This study uses one of the methods mentioned in the previous paragraph, lexical substitution, but instead of applying these randomly, we use a contextual embedding model to choose the most probable operations.  We demonstrate this technique with two samples below.  Each invocation of the algorithm could produce a novel alternative sample.

Original check-worthy tweet:
\begin{quote}
The country is panic stricken over the \#coronavirus, yet this flu season there have been 9.7 million flu illnesses, 87,000 hospitalizations and 4800 deaths from flu, including 32 pediatric deaths. It’s time to stop the panic and put things into perspective.
\end{quote}

Random Substitution 1: 
\begin{quote}
The country is panic tweeting over the \#coronavirus, yet this flu season there have been 9.7 million flu illnesses, 87,000 illnesses, 46 deaths from flu, including 32 pediatric deaths. It’s time you stop the panic and put things into perspective.
\end{quote}

Random Substitution 2:
\begin{quote}
The country remains panic stricken over the \#coronavirus, yet this flu pand there are been 9.7 million flu illnesses, 800,000 hospitalizations and 4800 deaths from flu, including 32 pediatric deaths. It’s time to stop the panic and put things into policy.
\end{quote}

Original check-worthy tweet: 
\begin{quote}
People aren’t surprised when I tell them there are 13,000 Covid-19 cases outside China, or when I tell them this number doubles every 3 days. But when I tell them that if growth continues at this rate, we’ll have 1.7 million cases in 3 weeks, they’re astonished.
\end{quote}

Random Substitution 1:
\begin{quote}
People aren’t surprised when I tell them there are 100,000 Covid-19 cases outside Minnesota, or when we tell them this number grows every 3 days. So when I tell them that if growth continues at this speed, we’ll have 1.7 million cases in 3 weeks, they’re astonished.
\end{quote}

Random Substitution 2
\begin{quote}
People aren't surprised when I tell her there are 13,000 X-19 cases outside China, than when I tell them this number doubled every 3 days. But when I tell them that if growth continues at my forecast, we'll have 1.7 million cases in 3 weeks, they're astonished.	
\end{quote}

More information on this procedure, including the models used, can be found in Section \ref{methoddataaugmentation}

\section{Exploratory Analysis}

Table \ref{tab1} shows the number of samples and unique word counts for each of the datasets provided.  We see that Arabic had the largest number of samples in training (3,095) while English had the least (688).  Similarly, Arabic had the highest count of unique words (29,619), and English had the lowest (4,499).  Assuming consistent data collection methodology and annotation standards across languages, we would hypothesize that larger datasets would yield higher-accuracy models.

\begin{table}
\centering
\caption{Dataset descriptions}\label{tab1}
\begin{tabular}{|l|l|c|c|}
\hline
Language & Modeling set & \# of samples & Unique word count\\
\hline
 & Train & 3,095& 26,919\\
 Arabic      & Test &  344& 5,413\\
       & Validation & 661& 8,242\\
\hline
 & Train&2,400 & 10,182\\
  Bulgarian     & Test & 600& 4,009\\
       & Validation & 350& 2,074\\
\hline
 & Train&698 & 4,499\\
  English     & Test & 124& 1,424\\
       & Validation & 140& 1,607\\
\hline
 & Train& 2,245 & 12,765\\
   Spanish    & Test &1,247 & 2,931\\
       & Validation & 250 & 8,744\\
\hline
 & Train& 1,709& 8,366\\
   Turkish    & Test & 1,90& 1,745\\
       & Validation & 388& 2,599\\
\hline
\end{tabular}
\end{table}

A qualitative analysis found that the topics included were consequential and that many tweets were worthy of fact-checking. Understanding the topics discussed in the data helps to evaluate the lexical coverage of our pre-trained models and helps to consider specialized pre-trained neural network models that could be relevant. \\

\textbf{Arabic.} The provided training set contains a large number of tweets referencing consequential political and human conflicts around the ``Houthis movement".  Notable portions of the training data also focus on diplomatic disagreements around the Algerian-Qatar relations. Political issues such as feminism are also discussed.  Finally, a smaller set of tweets reference COVID-19-related political events. The testing set contain mainly the latter two topics while validation set largely contained the former two.  Keywords that define these datasets include, but not limited to, ``Houthis", ``Yemen", ``Qatar",  ``Algerian", ``feminist", and "Veros Koruna".\\

\textbf{Bulgarian.} The training, testing, and validation sets all focus on COVID-19.  However, the tweets covers a rather varied set of topics, including political events, pandemic progression, and scientific and informational statements about the virus.  The most consistent keywords are amalgamations of the term ``COVID-19", including ``Koronavirus", ``Korona", ``Coronavius", and ``Covid".\\

\textbf{Spanish.} The conversations in the training, testing and validation sets are about political issues including: government corruption, unemployment, economic instability, the importance of education, political elections, and other related topics.  Spanish President, Pedro Sanchez, was named in a large number of tweets. Many tweets in the training set also mention climate issues, which are rare in the validation and testing sets.  Keywords that define these sets are ``President", ``Sanchez", ``government" and ``economy".  \\

\textbf{Turkish.} The training, testing, and validation set topics are varied but mostly related to Turkey domestic news, Turkish international affairs, and facts about Turkey as a country or the Turkish people.  The training contains a higher number of tweets covering Turkey's national/external debt, claims of mistreatment of children, the Syrian population, unemployment, and the Turkish economy. Some keywords include ``Turkey", ``Turkish", ``unemployment", ``President", ``Erdogan", ``presidential election" and ``Istanbul".\\

\textbf{English.} Tweets from all modeling sets are about COVID-19.  The training set contains many statements and information about COVID-19, global news about COVID-19, and COVID-19-related political decisions and events. The testing and validation set are more politically oriented with most tweets mentioning a political event.  The most frequent keywords include ``corona",``coronavirus" and other amalgamations of the term ``COVID-19".\\

\subsection{Label Balance}
All of the datasets provided by the CheckThat! organizers had label bias which skewed each dataset towards tweets that were not considered check-worthy.  The Turkish dataset had the highest percentage of check-worthy tweets (38\%), followed by English (35\%), Arabic (22\%), Bulgarian (13\%), and Spanish (8\%).  

\begin{figure}[h]
\caption{Label distribution across Arabic, Bulgarian, English, Spanish, and Turkish Training Sets}
\centering
\includegraphics[scale=0.8]{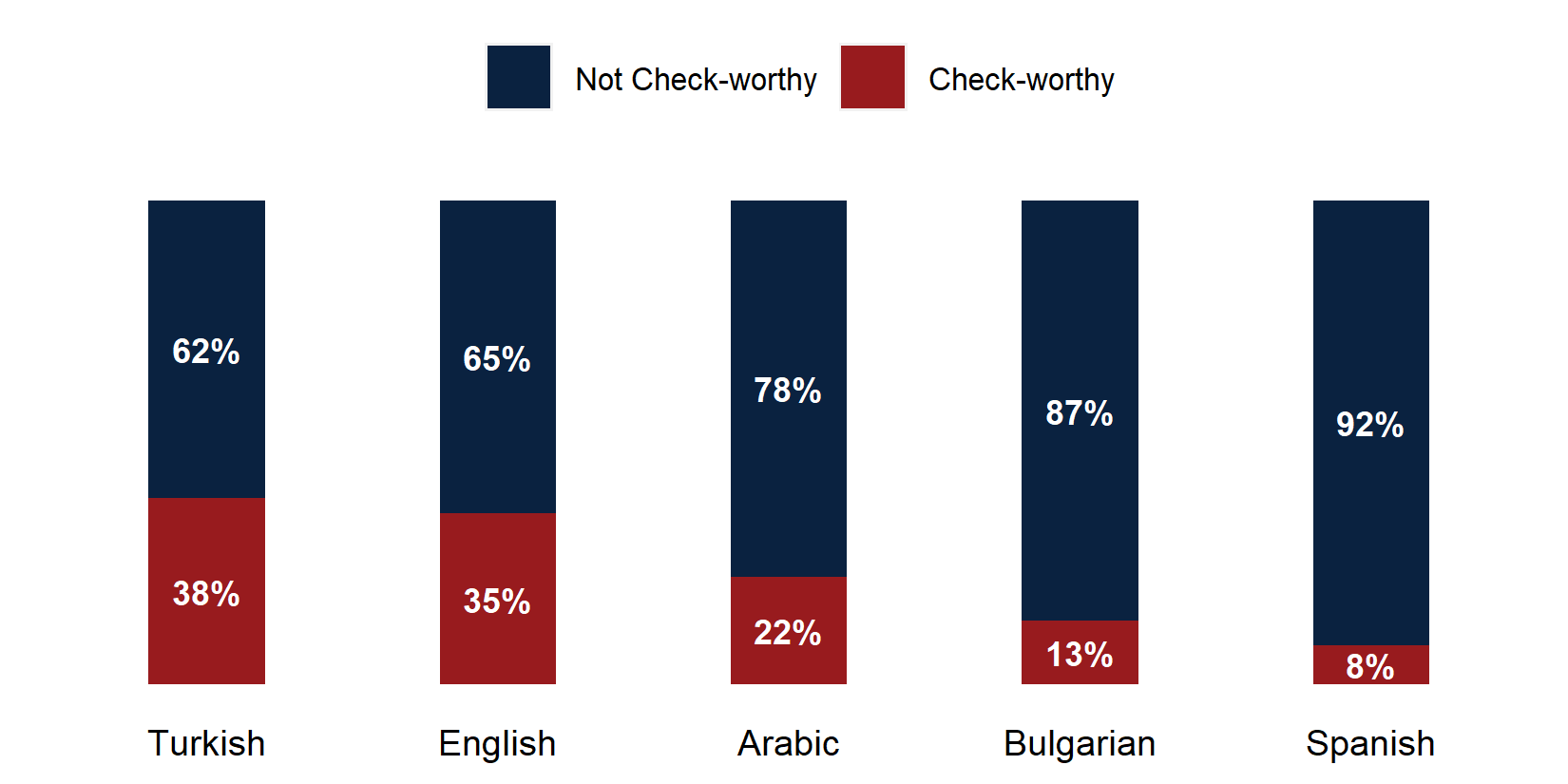}

\end{figure}

\subsection{WordPiece Analysis}

Transformer models utilize WordPiece tokenization schemes that are dependant on the model being evaluated.  At the time of pre-training, the WordPiece algorithm determines which pieces of words will be retained, and which will be discarded.  An UNK token is utilized as a placeholder in the lexicon, and used to represent WordPiece tokens received in novel input that did not get utilized at model creation.  We expect language samples which have a high amount of tokens processed as UNK would perform poorly. 

We present our analysis in Table \ref{unknown-token-distribution}. Most notably, Arabic training set contains over $120$K WordPieces, the largest number across all five languages, second by just over $110$K for Spanish. In addition, Arabic training set produced a much lower rate of unknown tokens ($0.291\%$) compared to Spanish ($2.313\%$).  Unexpectedly, the RoBERTa tokenizers we used did not return UNK tokens on any dataset provided by the CLEF CheckThat! organizers.

\begin{table}
\centering
\caption{Unknown token distribution in data for each language.}\label{unknown-token-distribution}
\begin{tabular}{|l|l|c|c|c|c|}
\hline
Language &  Tokenizer & Modeling Set & WordPiece & Unknown Token & Unknown Percent (\%)\\
\hline
 &  & Training & 122,991 & 358 & 0.291\\
Arabic  & BERT   & Testing & 14,184 & 35 & 0.247\\
  &   & Validation & 26,752 & 66 & 0.247\\
  \hline
&   & Training & 48,437  & 0 & 0\\
Bulgarian   & RoBERTa-based  & Testing & 12,172  & 0& 0\\
  &   & Validation & 5,799  & 0 & 0\\
  \hline
 &  & Training & 22,610  & 0  & 0\\
English  & RoBERTa-based & Testing & 3,704  & 0 & 0 \\
  &  & Validation & 4,856  & 0  & 0 \\
  \hline
 &  & Training &111,976 & 2,590 & 2.313 \\
 Spanish & BERT & Testing & 12,246 & 251 & 2.050 \\
  &  & Validation & 61,361 & 1,492 & 2.432 \\
  \hline
 &  & Training & 45,937 & 104 & 0.226 \\
 Turkish & BERT & Testing & 5,073 & 11 & 0.217 \\
  &  & Validation & 10,160 & 13 & 0.128 \\
\hline
\end{tabular}
\end{table}

\section{Transformer Architectures and Pre-trained Models}

\subsection{Pre-trained Models}

In this work, we utilize BERT and RoBERTa models.  The Bidirectional Encoder Representation Transformer (BERT) is a transformer-based architecture that was introduced in 2018 \cite{devlin2018bert,turc:2019}.  BERT has had a substantial impact on the field of NLP, and achieved state of the art results on 11 NLP benchmarks at the time of its release.  RoBERTa, introduced by \cite{roberta}, modified various parts of BERTs training process.  These modifications include more training data, more pre-training steps with bigger batches over more data, removing BERT's Next Sentence Prediction, training on longer sequences, and dynamically changing the masking pattern applied to the training data \cite{williams:ea:20}. 

For the Arabic Dataset, we used asafaya/arabic-bert-large \cite{safaya-etal-2020-kuisail}, which was trained on an Arabic version of OSCAR, an Arabic Wikipedia dump, and other Arabic resources.  It contains a vocabulary of length of 32,000.

For Turkish and Spanish, we used dbmdz/bert-base-turkish-cased \cite{stefan_schweter} and geotrend/bert-base-es-cased \cite{smallermbert} respectively.  The Turkish BERT model contains a vocabulary of length 32,000 and the Turkish model contains a vocabulary of length 26,359.

For English and for Bulgarian, we used roberta-large \cite{DBLP:journals/corr/abs-1907-11692} and iarfmoose/roberta-base-bulgarian \cite{bulgarian_roberta} respectively.  The English RoBERTa model contains 50,265 WordPieces, and the Bulgarian RoBERTa model contains 52,000 WordPieces.

\section{Method}

\subsection{Data Augmentation}
\label{methoddataaugmentation}

The organizers provided a training and a development set for each language. We created 80/20 stratified splits on each training set to create internal training and validation sets for experimentation.  We used the development set provided by organizers as a hold-out test set.  For each of the internal training datasets, we extracted positive labels and performed contextual word embedding augmentation on each of the positive labels, one epoch at a time, repeating until the number of positive labels at each epoch exceeded the number of negative labels.

For each language, augmentation and training were done with BERT or RoBERTa models.  BERT-based contextual embedding models were used for Arabic \cite{safaya-etal-2020-kuisail}, Spanish \cite{smallermbert}, and Turkish \cite{stefan_schweter}, and RoBERTa-based contextual embedding models were used for English \cite{DBLP:journals/corr/abs-1907-11692} and Bulgarian \cite{bulgarian_roberta}.  We used \cite{ma2019nlpaug}\footnote{https://github.com/makcedward/nlpaug} to apply Contextual Word Embedding Augmentation.  This augmentation type uses the surrounding words of a tweet to apply the most probable insertion or substitution of a lexical item. We chose to only apply substitution.

With Contextual Word Embedding Augmentation, the user must determine the probability a token should be augmented.  On the Bulgarian and Turkish datasets, we tried using no augmentation (p = null) and augmentation at p = \{0.1, 0.2, 0.3, 0.4, 0.5\}.  Additionally, we explored back-translation using AWS translation. We appended back-translated check-worthy tweets to the training set, using English as a pivot language.  In the table below, we show the precision, recall, and f1 score for check-worthy class on the Bulgarian dataset.  The Turkish experiments yielded similar results.  For both languages, we found that augmentations at p = 0.1 resulted in a significant increase in recall and f1 for check-worthy tweets, as we show in the table below. Augmentations at higher probability thresh-holds also yielded better recall and f1 than our null model, but not better than at p = 0.1.  We found that augmentation at p=0.1 provided better precision, recall and f1 than using back-translation (translation). Due to time and cost limitations, we did not repeat back-translation experiments for other competition languages or using other pivot languages.

\begin{table*}[]
  \caption{Augmentation substitution improvements on Bulgarian training set using varied token augmentation probabilities}
  \label{tab:augprobs}
  \centering
    \begin{tabular}{llllllll}
                  & \textbf{p=null} & \textbf{p=0.1} & \textbf{p=0.2} & \textbf{p=0.3} & \textbf{p=0.4} & \textbf{p=0.5} & \textbf{translation} \\
    cw\_precision & 0.65   & 0.51  & 0.41  & 0.36  & 0.36  & 0.42  & 0.47        \\
    cw\_recall    & 0.17   & 0.6   & 0.66  & 0.45  & 0.31  & 0.31  & 0.47        \\
    cw\_f1        & 0.28   & 0.55  & 0.51  & 0.4   & 0.33  & 0.36  & 0.47       
    \end{tabular}
\end{table*}

Based on this exploration (and due to the computational cost of this technique) we adopted p = 0.1 for all languages.

For each language, BERT- or RoBERTa-augmented check-worthy examples were appended to the training data and given a check-worthy label.  In all languages, we found this improved f1 score for the check-worthy class when applied to our hold-out dev set.  However, it's possible that this technique limited some  models' performance on the evaluation test set. The table below displays the number of check-worthy augmented samples that were generated for each language.

\begin{table*}[]
    \caption{Number of augmented check-worthy samples generated for each language}
    \begin{tabular}{llllll}
    \textbf{Language}             & Bulgarian  & Arabic  & English  & Spanish  & Turkish \\
    \textbf{Augmentations} & 2,471     & 2,748  & 492     & 1,800   & 656    
    \end{tabular}
\end{table*}

\subsection{Classification}
For both BERT and RoBERTa, we added an additional mean-pooling layer and dropout layer on top of the model prior to the final classification layer.  Adding these additional layers has been shown to help prevent over-fitting while fine-tuning \cite{williams:ea:20}.  We used an Adam optimizer with a learning rate of 1.5e-5 and an epsilon of 1e-8. We use a binary cross-entropy loss function, 2 epochs, and a batch size of 32.

\subsection{Ranking}
To generate rankings, the model's outputs were fed through a SoftMax function.  The difference between the positive and negative class likelihoods were then used to rank tweets.

\section{Results}

The official metric of the Lab was mAP for all languages.  Table \ref{tab:accentureresults} lists our results.  Arabic performed the best with 0.658.  This was followed by Bulgarian (0.497), Spanish (0.491), Turkish (0.402), and English (0.101).  

\begin{table*}[]
  \caption{Accenture results from CheckThat! 2021 Task 1 }
  \label{tab:accentureresults}
\begin{tabular}{l
l llllllll}
\textbf{Entry} & \textbf{mAP}   & \textbf{MRR}  & \textbf{RP}               & \textbf{P@1}               & \textbf{P@3}& \textbf{P@5}& \textbf{P@10}& \textbf{P@20}& \textbf{P@30} \\

Arabic & \multicolumn{1}{r}{
0.658} & \multicolumn{1}{r}{1.000} & \multicolumn{1}{r}{0.599} & \multicolumn{1}{r}{1.000} & \multicolumn{1}{r}{1.000} & \multicolumn{1}{r}{1.000} & \multicolumn{1}{r}{1.000} & \multicolumn{1}{r}{0.9500} & \multicolumn{1}{r}{0.840}\\
Bulgarian     & \multicolumn{1}{r}{
0.497} & \multicolumn{1}{r}{1.000} & \multicolumn{1}{r}{0.474} & \multicolumn{1}{r}{1.000} & \multicolumn{1}{r}{1.000} & \multicolumn{1}{r}{0.800} & \multicolumn{1}{r}{0.700} & \multicolumn{1}{r}{0.600} & \multicolumn{1}{r}{0.440}\\
English     & \multicolumn{1}{r}{
0.101} & \multicolumn{1}{r}{0.143} & \multicolumn{1}{r}{0.158} & \multicolumn{1}{r}{0.000} & \multicolumn{1}{r}{0.000} & \multicolumn{1}{r}{0.000} & \multicolumn{1}{r}{0.200} & \multicolumn{1}{r}{0.200} & \multicolumn{1}{r}{0.100}\\
Spanish     & \multicolumn{1}{r}{
0.491} & \multicolumn{1}{r}{1.0000} & \multicolumn{1}{r}{0.508} & \multicolumn{1}{r}{1.0000} & \multicolumn{1}{r}{0.667} & \multicolumn{1}{r}{0.800} & \multicolumn{1}{r}{0.900} & \multicolumn{1}{r}{0.700} & \multicolumn{1}{r}{0.620}\\
Turkish     & \multicolumn{1}{r}{
0.402} & \multicolumn{1}{r}{0.250} & \multicolumn{1}{r}{0.415} & \multicolumn{1}{r}{0.000} & \multicolumn{1}{r}{0.000} & \multicolumn{1}{r}{0.400} & \multicolumn{1}{r}{0.400} & \multicolumn{1}{r}{0.650} & \multicolumn{1}{r}{0.660}\\

\end{tabular}
\end{table*}

The results from Arabic and English can be compared to Accenture's results in CheckThat! 2020.  Compared to last year, our team's Arabic score increased and our team's English score decreased.  

\section{Discussion}
In Section \ref{methoddataaugmentation}, we showed that contextual embedding augmentation on top of the current training data improved the f1 score of our systems.  In the Lab ranking, however, we found that the Accenture received the top results for Arabic, and performed less well for other languages.  Table \ref{tab:mapresults_trainingsamplecount} shows our mean average precision versus the number of training samples provided by the CheckThat! organizers.

\begin{table}[]
 \caption{Accenture mAP results from CheckThat! 2021 Task 1, with training sample count }
  \label{tab:mapresults_trainingsamplecount}
\begin{tabular}{lll}
\textbf{}          & \textbf{mAP} & \textbf{\# Samples} \\
\textbf{English}   & 0.101        & 698                 \\
\textbf{Turkish}   & 0.402        & 3095                \\
\textbf{Spanish}   & 0.491        & 2245                \\
\textbf{Bulgarian} & 0.497        & 2400                \\
\textbf{Arabic}    & 0.658        & 3095               
\end{tabular}
\end{table}

We believe the transformer methods we employed to be highly sensitive to the quantity of training data and distribution of topics across the split dataset.  While we were able to generate additional training data, and improve the results, we believe even more augmentation should be employed where natural labeled text cannot be acquired.

Back translation, which we employed last year, worked well for this problem.  We employed it only on Arabic, but expect the technique would work well for the other languages in the Lab as well.  We used Contextual Word Embedding Augmentation, this year, but limited our transformations to swaps.  Lexical insertion may show to be useful as well.  Synonyms and hypernym replacement would likely show advantage.

Because of the computational cost of augmenting the data with contextual embedding augmentation, we chose p = 0.1 for all languages.  The optimal value is likely to be language and task dependant. We would recommend a parameter search of this value.

\section{Conclusion}
This paper presents results from the Accenture Team for CLEF2021 CheckThat! Lab, Task 1, to analyze English, Arabic, Spanish, Turkish, and Bulgarian social media to identify claims that require fact-checking.  We presented a methodology that provided NLP augmentation of the training data to create additional synthetic training samples.   We found this method improved our results.  This approach received the highest mean average precision in the Lab this year for Arabic. 

\bibliographystyle{splncs04}
\bibliography{clef2021-checkthat-paper}

\end{document}